\documentclass[sigconf, nonacm]{acmart}

\usepackage{threeparttable}
\usepackage{balance}

\AtBeginDocument{%
  \providecommand\BibTeX{{%
    \normalfont B\kern-0.5em{\scshape i\kern-0.25em b}\kern-0.8em\TeX}}}

\setcopyright{none}

\usepackage[utf8]{inputenc} 
\usepackage[T1]{fontenc}    
\usepackage{hyperref}       
\usepackage{url}            
\usepackage{booktabs}       
\usepackage{amsfonts}       
\usepackage{nicefrac}       
\usepackage{microtype}      
\usepackage{amsmath}
\usepackage{graphicx}
\usepackage{algorithmic}
\usepackage{algorithm}
\usepackage{xcolor}
\usepackage{multirow}
\usepackage{subcaption}
\usepackage{enumitem}
\numberwithin{algorithm}{section}
\usepackage[export]{adjustbox}

\usepackage{letters}
\usepackage{operators}

\definecolor{sred}{RGB}{255,63,88}
\definecolor{mred}{RGB}{191,47,66}
\definecolor{lred}{RGB}{122,30,42}

\title{Value Function is All You Need: A Unified Learning Framework for Ride Hailing Platforms}

\usepackage[belowskip=2pt,aboveskip=2pt]{caption}

\begin{document}
\author{
Xiaocheng Tang\textsuperscript{1},
Fan Zhang\textsuperscript{1},
Zhiwei (Tony) Qin\textsuperscript{1},
Yansheng Wang\textsuperscript{2},
Dingyuan Shi\textsuperscript{2},
Bingchen Song\textsuperscript{2},
Yongxin Tong\textsuperscript{2},
Hongtu Zhu\textsuperscript{1},
Jieping Ye\textsuperscript{3}
}
 \affiliation{%
  \institution{\textsuperscript{1}AI Labs, Didi Chuxing}
  \country{}
 }
 \email{{xiaochengtang, feynmanzhangfan, qinzhiwei, zhuhongtu}@didiglobal.com}
 \affiliation{%
  \institution{\textsuperscript{2}School of Computer Science and Engineering and IRI, Beihang University, China}
  \country{}
 }
 \email{{arthur_wang, chnsdy, songbch, yxtong}@buaa.edu.cn}
 \affiliation{%
  \institution{\textsuperscript{3}University of Michigan, Ann Arbor, United States}
  \country{}
 }
 \email{jpye@umich.edu}


%



\begin{abstract}
Large ride-hailing platforms, such as DiDi, Uber and Lyft, connect tens of thousands of vehicles in a city to millions of ride demands throughout the day, providing great promises for improving transportation efficiency through the tasks of order dispatching and vehicle repositioning.
Existing studies, however, usually consider the two tasks in simplified settings that hardly address the complex interactions between the two, the real-time fluctuations between supply and demand, and the necessary coordinations due to the large-scale nature of the problem.
In this paper we propose a unified value-based dynamic learning framework (V1D3) for tackling both tasks.
At the center of the framework is a globally shared value function that is updated continuously using online experiences generated from real-time platform transactions.
To improve the sample-efficiency and the robustness, we further propose a novel periodic ensemble method combining the fast online learning with a large-scale offline training scheme that leverages the abundant historical driver trajectory data.
This allows the proposed framework to adapt quickly to the highly dynamic environment, to generalize robustly to recurrent patterns and to drive implicit coordinations among the population of managed vehicles.
Extensive experiments based on real-world datasets show considerably improvements over other recently proposed methods on both tasks. Particularly, V1D3 outperforms the first prize winners of both dispatching and repositioning tracks in the KDD Cup 2020 RL competition, achieving state-of-the-art results on improving both total driver income and user experience related metrics.
\end{abstract}
\maketitle

\begin{figure}[h]
\begin{center}
    \hspace*{-0.0in}\adjincludegraphics[width=0.48\textwidth, trim={0 0 0 0}, clip]{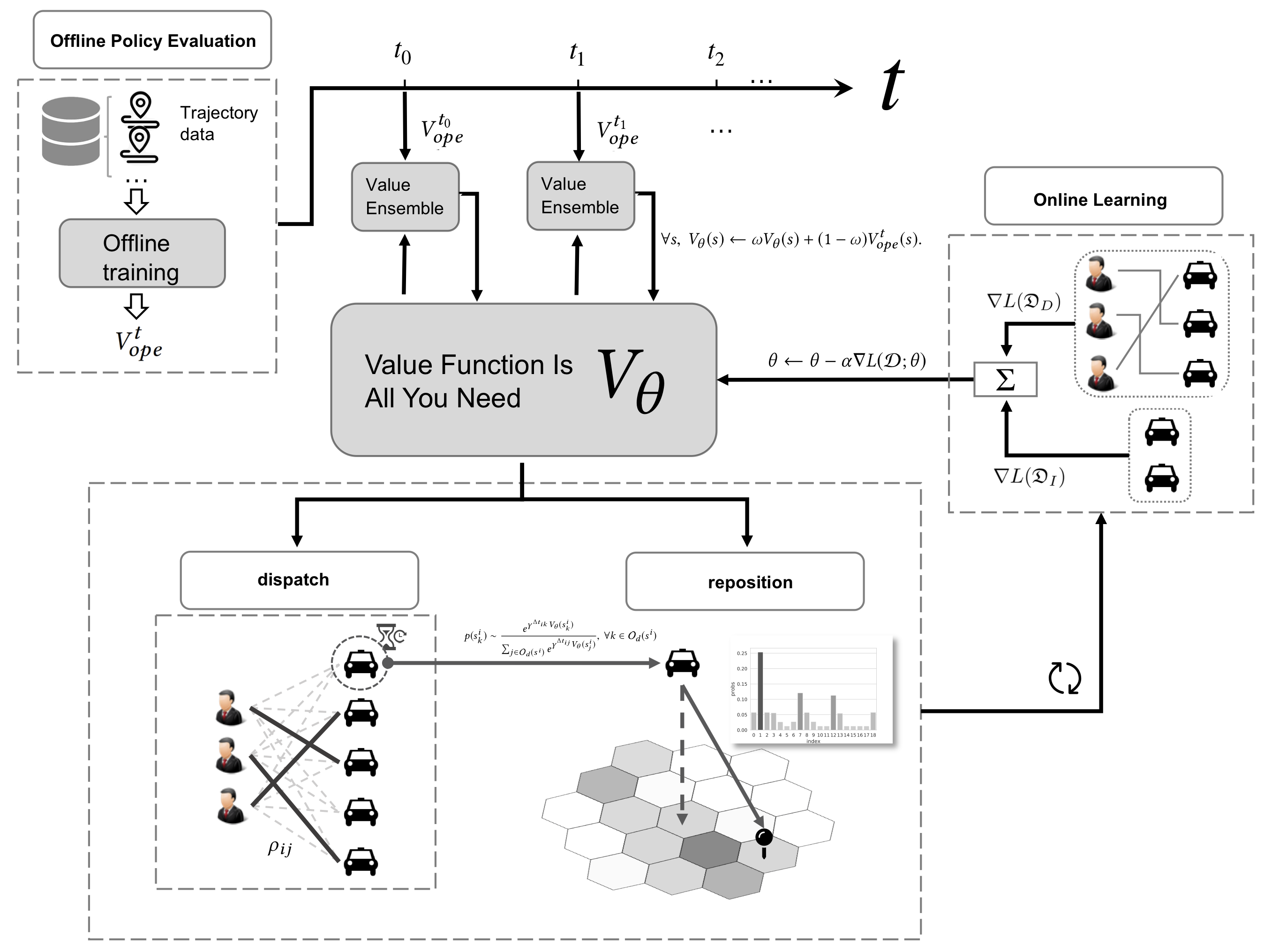}
    \caption{An illustration of V1D3. At the center is the globally shared value function that is continuously updated by both online learning and value ensemble from the offline model.}
    \label{fig:cover}
\end{center}
\vspace{-5pt}
\end{figure}

\section{Introduction}

Popular online ride-hailing platforms, such as Uber, Lyft, and DiDi, have revolutionized the way people travel and socialize in cities across the world and are increasingly becoming essential components of the modern transit infrastructure.
The rising prevalence of these ride-hailing platforms,
seamlessly connecting passengers with drivers through their smart mobile phones,
has provided great promises for the optimization of transportation resources with the direct access to the abundant amount of real-time transit information.
This new possibility has motivated a surge of interest at emerging academic research and algorithm development (see, e.g., \cite{qin:wagner2020,shou2019find,tlab2021,ozkan2017dynamic}), centering around one of the core questions --
how to leverage the real-time rich information to reduce the inefficiencies from limited supply (drivers) and asymmetric demand (passengers) across time and space in a highly dynamic environment such as the ride-hailing marketplaces. Highly efficient marketplaces present significant benefits to drivers and passengers through improved income and reduced waiting time, among a number of metrics.

A large portion of this work aims to improve the performance of the platforms by considering the optimization of the two main operational functions of ride-hailing systems, order dispatching and vehicle repositioning, both of which have great influences on the balance between supply and demand conditions.
Order dispatching \cite{tang2019vnet,wang2018deep,xu2018large,ozkan2017dynamic,korolko2018dynamic} matches idle drivers (vehicles) to open trip orders, transporting passengers (and the drivers) to the trip destinations.
Intuitively, the decision of one dispatching action directly impacts the future spatial distributions of the drivers which in turn play a role in determining the outcome of the future dispatching decision. A better alignment between
the distribution of drivers and that of the orders, for example, will result in an decrease in the passenger waiting time, and consequently an improvement of the order completion rate as well as the passenger satisfactions.
Vehicle repositioning \cite{shou2019find,lin2018efficient,oda2018movi,tlab2021}, on the other hand,
aims to improve the platforms' performance
by deploying idle vehicles to a specific location in anticipation of future demand at the destination or beyond.
Compared to dispatching where the vehicle transitions are determined by the origins and destinations of the outstanding orders, the repositioning operation is a proactive measure that has the freedom to make recommendations of any feasible destinations inside the city.
On the flip side,
it becomes a much complicated problem when the goal is to improve the average driver income of a large fleet for an extended period of time (e.g., more than 8 hours per day), which often requires some form of coordination among the vehicles to avoid causing undesirable competitions, e.g., crowding too many vehicles into a single high-demand location.


The sequential nature of both problems has motivated much of the work in this field to optimize the long-term reward under the settings of reinforcement learning (RL).
Indeed one of the most recent works CVNet \cite{tang2019vnet} successfully applies a deep reinforcement learning approach to the problem of order dispatching and demonstrates significant improvement on both total driver income and user experience related metrics through real-world online experiments. In particular, the work estimates a spatio-temporal value function representing drivers' future return given the current state from the historical driver trajectories using an offline policy evaluation method.
During online planning the value function is used to populate the edge weights of a bipartite matching problem such that the objective is to maximize the total return of all drivers.
Crucially, the work highlights the importance of
handling the real-time dynamics between demand and supply in an online environment.
In light of this, CVNet augments the historical training data with contextual information which is fed into the value network as additional inputs.
Relying simply on contextual inputs with a model trained on historical data to reflect the real-time supply/demand conditions, however, can be too restrictive and difficult to work in practice. For one thing, this requires the neural network to not only generalize well to real-time contextual conditions, which often endure high variances and irregularities, but also maintain robustness with respect to the main state inputs as well.

Our contribution in this paper is a unified \textbf{V}alue-based \textbf{D}ynamic learning framework for both order \textbf{D}ispatching and \textbf{D}river repositioning (\textbf{V1D3}) that adapts quickly to the variations of real-time dynamics and that maintains high performance scaling to the real-world scenarios involving tens of thousands of agents (vehicles).
Extensive experiments within DiDi's real data-backed ride-hailing simulation environment, which hosted the KDD Cup 2020 RL track competition\footnote{The competition submission platform is open for public and can be accessed at \url{https://outreach.didichuxing.com/Simulation/}},
demonstrate that the proposed method significantly outperforms the state-of-the-art methods including the \emph{first prize} winner of the order dispatching track and that of the vehicle repositioning track in KDD Cup 2020.
There are significant challenges for developing a high-performing framework for both tasks that scales to the large-scale real-world scenarios.
First of all, the interactions between the two tasks need to be properly accounted for in order to achieve the best performance.
The repositioning task, for example, needs to recognize the "dispatching probability" which varies with the dispatching policy, while the dispatching task has to take into account the repositioning results that change the supply distributions.
Relying simply on models with contextual dependencies and real-time inputs but only trained offline has practical limitations for reasons mentioned above, let alone having multiple such models interact with each other in real time.
Secondly, the large-scale setting of both tasks requires efficient coordinations among agents both to resolve dispatching constraints and to avoid undesirable competitions among vehicles as mentioned above. A straightforward application of multi-agent DRL \cite{Tampuu2017,Lowe2017}, for example, only allow coordinations among a small set of agents due to prohibitive computational costs.
Finally, there are not only daily variations of supply and demand in the ride-hailing marketplaces that both tasks need to handle, but also the occurrences of irregular events, such as a large musical concert which can result in a dramatic fluctuations of the demands during a short period of time around a local area.

Interestingly we find in this work that all the challenges above can be well addressed by building upon the value-based approaches originally proposed only for order dispatching \cite{tang2019vnet,xu2018large}.
We show that a global centralized value function is all you need for both dispatching and repositioning in a large-scale setting involving tens of thousands of or more drivers.
In particular we propose a novel population-based online learning objective, derived from the on-policy value iteration \cite{sutton1998reinforcement}, which trains and updates the value function continuously during online execution using real-time transactions between drivers and orders.
Updating the centralized value function through this objective can be considered as minimizing the population mean of squared Temporal Difference (TD) error from each new driver transition trajectory resulted from the current dispatching action.
Updates are performed within each dispatching window, where new transitions take place from either idle movements or trip assignments. This allows the value function to adapt quickly to any new changes in the environment.

There are, however, important limitations for
relying solely on online learning without learning from the history.
Its sample-inefficiency due to the need for infinite online exploration is, of course, well known \cite{Singh:2000ur}.
More importantly, the ride-hailing environment is a time-varying system with multiple systematic shifts of the state distribution and dynamics throughout the day, e.g., the transition from morning rush hours to off-peak hours around noon.
Note, however, that this systematic distributional shift follows strong patterns that can be more easily captured in the historical data, compared to the aforementioned irregular events and daily variations in supply and demand which appear more like random noise that are hard to learn by offline training alone.


To address the above limitations, we further propose a novel ensemble method combining the fast online learning with a large-scale offline policy evaluation scheme building upon the previous work CVNet \cite{tang2019vnet} that leverages the abundant historical driver trajectory data.
The proposed method maintains a centralized value function, adapted quickly with new online experiences, while periodically ensembled with the offline-trained value function at \emph{the corresponding temporal slice}.
Together we show that the value function not only is able to capture general time varying patterns and generalizes across the history of episodes, but also accounts for the noisy variations of the \emph{current} episode through the online learning procedures.
Crucial to this capability is that the value function behaves as a "shared memory" between the two tasks, refreshed whenever new changes take place, such that by acting through the "shared memory" repositioning recognizes what to expect from dispatching while dispatching is aware of the latest changes resulted from repositioning.
Finally, the updates applied to the value function operate as an implicit form of coordinations among the controlled agents, with a "feedback loop" such that previous repositioning results are communicated in the global value store which informs the current decision making accordingly.

The rest of the paper is organized as follows. We start by laying out the notations and definitions used in the paper in Section \ref{sec:Preliminaries}. Then we derive the online learning objective in Section \ref{sec:population_based_online_learning_objective} and describe the periodic ensemble method in Section \ref{sec:offline}. The unified framework for dispatching and repositioning based on the value function is detailed in Section \ref{sec:unified_framework_for_dispatch_and_reposition} along with the complete algorithmic procedures. Experiment results are presented in Section \ref{sec:experiments}. We conclude our work in Section \ref{sec:conlude}.




\section{Preliminaries} 
\label{sec:Preliminaries}

We consider the activities of each driver as a \emph{semi-Markov decision process} (SMDP) \cite{tang2019vnet} with a set of temporally extended actions known as \emph{options}.
At decision point $t$ the driver $i$ takes an option $o_t^i$, transitioning from current state $s_t^i$ to the next state $s_{t'}^i$ at time step $t'$ while receiving a numerical reward $r_t^i$.
The available options to take at each state can be either a trip assignment or an idle reposition, which can be compactly represented as a tuple consisting of the destination and the reward, i.e., $(s_{t'}^i, r_t^i)$ where the duration of the transition is given by $t' - t$.
The reward is equal to the trip fee if the option is a trip assignment, and is zero if the option is an idle reposition.
The driver enters the system and starts taking order requests at the start of the day $t=0$, and finishes the day, or the episode, at the \emph{terminal} time step $t=T$. Through the episode a policy $\pi(o_t|s_t)$, or $\pi_t$, specifies the probability of taking option $o_t$ in state $s_t$. The state value function $V^{\pi}(s) := E\{\sum_{j=t+1}^T \gamma^{j-t-1} r_j | s_t = s\}$ for the policy $\pi$, is the expected long-term discounted reward obtained at a state $s$ and following $\pi$ thereafter till the end of the episode. $V^{\pi}(s)$ is the fixed point of the Bellman operator $\Tcal^{\pi}$, $\forall s$: $\Tcal^{\pi} V(s) := E \{r_{st}^o + \gamma^{t'-t} V^{\pi} (s_{t'}) | s_t = s\}$ where $r_{st}^o$ is the corresponding accumulative discounted reward received through the course of the option \cite{tang2019vnet}.




\section{Population-based Online Learning Objective} 
\label{sec:population_based_online_learning_objective}

In this section, we present the online method to learn and update the value function such that it captures the nonstationary dynamics of the supply-demand conditions in real time.
Later we will discuss how to complement it with offline training the value function in the unified optimization framework for both dispatching and repositioning.

Consider the set of available drivers $\Dcal$ in the current dispatching window.
After dispatching the states of the drivers will change based on different options the drivers execute.
Hence we update the value functions accordingly accounting for each driver's different transition. In particular,
let the set $\Dcal_D$ denote the set of drivers successfully assigned with orders. Let $\Dcal_I := \Dcal \setminus \Dcal_D$ denote the idle drivers that have not been assigned with orders in the current dispatching round.
For each driver $i \in \Dcal_D$, let $s^i_{driver}$ denote the current driver state and $s^i_{order}$  denote the destination state of the order assigned to the driver $i$. The one-step Bellman update for this transition is then given by,
\begin{align}
\label{equ:v-pos}
  V(s^i_{driver}) \leftarrow r^i_{order} + \gamma^{\Delta t_{order}} V(s^i_{order})
\end{align}
where $r^i_{order}$ is the corresponding order trip fee and $\Delta t_{order}$ is the estimated order trip duration.

For each driver $i \in \Dcal_I$, let $s^i_{idle}$ denote the next state after idle movement from the current state $s^i_{driver}$. Then the Bellman update for this idle transition is given by,
\begin{align}
\label{equ:v-neg}
  V(s^i_{driver}) \leftarrow 0 + \gamma^{\Delta t_{idle}} V(s^i_{idle})
\end{align}
where the transition yields 0 reward and lasts for $\Delta t_{idle}$ duration.
Following practical Q-learning methods (e.g., \cite{van2016deep,mnih2015human}), we can convert the above Bellman updates into a bootstrapping-based objective for training a V-network, $V_{\theta}$, via gradient descent. This objective is also known as mean-squared temporal difference (TD) error. Particularly, let $\delta^i_{\theta}$ represent the TD error for the $i$th driver and we obtain,
\begin{align}
\label{equ:delta-td-error}
\delta^i_{\theta} = \left\{
\begin{array}{ll}
r^i_{order} + \gamma^{\Delta t_{order}} V_{\theta}(s^i_{order}) - V_{\theta}(s^i_{driver})
& \forall i \in \Dcal_D;\\
\gamma^{\Delta t_{idle}} V_{\theta}(s^i_{idle}) - V_{\theta}(s^i_{driver})
& \forall i \in \Dcal_I.
\end{array} \right.
\end{align}
Applying it to all drivers in $\Dcal$, we obtain the \emph{population-based mean-squared TD error}:
\begin{align}
\label{equ:pop-td-error}
  \notag \min_{\theta} ~L(\Dcal;\theta) :=
  \notag &\sum_{i \in \Dcal_D} (V_{\theta}(s^i_{driver}) - r^i_{order} - \gamma^{\Delta t_{order}} \bar V_{\theta}(s^i_{order}))^2 \\
  + &\sum_{i \in \Dcal_I} (V_{\theta}(s^i_{driver}) - \gamma^{\Delta t_{idle}} \bar V_{\theta}(s^i_{idle}))^2 = \sum_{i \in \Dcal} (\delta^i_{\theta})^2
\end{align}
Here following the common practice \cite{van2016deep} a target network $\bar V_{\theta}$ is used to stabilize the training, which acts as a delayed copy of the V-network $V_{\theta}$.
After each round of dispatch we update $V_{\theta}$ by taking a gradient descent step towards minimizing $L(\theta)$, i.e.,
$\theta \leftarrow \theta - \alpha \nabla L(\Dcal;\theta)$
where $\alpha > 0$ is a step-size parameter to control the learning rate.


\section{Value Ensemble with Offline policy Evaluation} 
\label{sec:offline}
Online learning enables fast adaptations in real time, but suffers from both sample-inefficiency and the overemphasis on "recency" while overlooking important global patterns which can be more easily captured by learning from the large offline datasets.
In fact,
from \eqref{equ:pop-td-error} it can be seen that the size of the online training data depends on the number of drivers and their corresponding states in the system.
Hence the effectiveness of online learning can vary noticeably depending on the availability of drivers and the scale of the operating city
given that the performance of RL methods usually hinges on a sufficient coverage of state distribution in the training data.
In this section we discuss the remedies to these issues and propose a periodic ensemble method to incorporate the knowledge from offline training methods.

\subsection{Regularized Offline Policy Evaluation} 
\label{sub:regularized_off_policy_evaluation}

We adopt the approach proposed in \cite{tang2019vnet} for estimating the state value function from the historical driver trajectories $\Hcal$.
The objective of offline policy evaluation (OPE) is obtained by applying the Bellman squared error to each driver transition extracted from the full trajectory of each episode.
Each transition is represented by a tuple $(s, R, s') \in \Hcal$ meaning that the driver moves from state $s$ to $s'$ while receiving a reward of $R$. Given such transition dataset $\Hcal$ the learning objective can be obtained as follows,
\begin{align}
\label{equ:ope-vnet}
    \notag &\min_{\rho} ~L_{ope}(\Hcal; \rho) := \\
    &E_{(s, R, s') \sim \Hcal}\left[ (R + \gamma^{\Delta t} \hat V_{ope}(s', t' | \rho) - V_{ope}(s, t|\rho))^2 \right] + \lambda \cdot L_{reg}(\rho)
\end{align}
where $R$ denotes the properly discounted reward from the transition based on the Semi-MDP formulation \cite{tang2019vnet}, $\hat V_{ope}$ is the target network \cite{van2016deep} and the regularization term
$L_{reg}$ is added to induce a smooth and robust value response by minimizing an upper bound on the global Lipschitz constant of the neural network $V_{ope}(\cdot | \rho)$ with trainable weights $\rho$.
Note that to account for the time-varying aspect of the system,
we augment the input state to the value function with the current time stamp, which is one main difference from the online objective \eqref{equ:pop-td-error}.
This allows us to obtain a time series of state value functions which will be used as the basis for ensemble with the online value function. More implementation details can be found in the Appendix \ref{apx:ope}.

\subsection{Periodic Value Ensemble} 
\label{sub:periodic_value_ensemble}

In the online environment we maintain and update $V_{\theta}$
using the results of each dispatching round according to \eqref{equ:pop-td-error}.
To account for the non-stationarity of the environment, we periodically `reinitialize' $V_{\theta}$ with a weighted ensemble scheme combining the latest state of $V_{\theta}$ and the snapshot of the offline trained $V_{ope}^t$. Specifically, let $\Ecal$ denote the set consisting of changing time points when the re-ensemble is triggered. At the current time step if $t \in \Ecal$, then we re-ensemble as follows,
\begin{align}
\label{equ:weighted_ensemble}
    \forall s,  ~
    V_{\theta}(s) \leftarrow
    \omega V_{\theta}(s) + (1 - \omega) V_{ope}^t(s).
\end{align}
where $\omega > 0$ is a hyperparameter to balance the weighting between online value function and offline trained values. Note that $V_{ope}$ is trained with the current time stamp as part of the input as in \eqref{equ:ope-vnet} such that $V_{ope}^t(s)$ can be obtained by fixing time at $t$ for each state $s$.
The set $\Ecal$ can be determined by learning a segmentation on the historical aggregated order time series to identify the temporal boundaries of the order distributional shift \cite{TRUONG2020107299} (for implementation details refer to Appendix \ref{apx:ecal}).


There are important nuances in learning mechanisms between $V_{\theta}$ and $V_{ope}$. To better understand the intuition behind \eqref{equ:weighted_ensemble}, note that while the full driver trajectory is known and available for learning the offline value $V_{ope}$, for $V_{\theta}$ only the partial driver trajectory is available for training since it is updated online in a temporally sequential order. At time step $t$, for example, $V_{ope}^t$ reflects the historical trajectories from $t$ till the end of the episode, while $V_{\theta}$ is trained on the trajectories from the beginning of the \emph{current} episode till time $t$.
Through weighted ensemble of both $V_{ope}^t$ and $V_{\theta}$, we are able to capture general time varying patterns across the history of episodes while also accounting for individual variations of the \emph{current} episode.




\section{Unified Framework For Dispatch and Reposition} 
\label{sec:unified_framework_for_dispatch_and_reposition}
We now describe the method to dispatch orders and to reposition vehicles based on the value function $V_{\theta}$.
The dispatch method adopts the approach used in \cite{xu2018large,tang2019vnet} which embeds the value function into a combinatorial problem to resolve the dispatching constraints in real time. It can be seen as a policy improvement step \cite{sutton1998reinforcement} in a multi-agent environment with thousands of drivers.
Reposition shares the same centralized value function with dispatching and computes the action in a value-based probabilistic manner.
We will demonstrate empirically later that this simple approach, when integrated into the proposed unified framework, can achieve robust and superior performance
even when the number of managed vehicles scales to a significant portion of the whole vehicle population on the platform.


\subsection{Planning With Multi-driver Dispatching}\label{sub:dispatch}
The order-dispatching system of the ride-hailing platforms can be viewed as a centralized planner continuously querying and processing the system state involving the current outstanding requests and available vehicles.
Within each such decision window a bipartite matching problem is formulated based on the current system state. The solution of the problem makes sure that dispatching constraints are respected while the sum of utility scores is maximized. Mathematically it can be written as a constrained optimization with the utility score $\rho_{ij}$ indicating the value of matching each drive\textbf{}r $i$ to an order $j$, as follows:
\begin{eqnarray}
\label{equ:dispatch-obj}
\text{arg}\max_{x_{ij}} \sum_{j=0}^M \sum_{i=0}^N \rho_{ij} x_{ij},
\text{  s.t.}  \,\, \sum_{j=0}^M x_{ij} \leq1 \,\, \forall i; \,\, \sum_{i=0}^N x_{ij} \leq1\,\, \forall j.
\end{eqnarray}
where
\[ x_{ij} = \left\{ \begin{array}{ll}
1 & \mbox{if order $j$ is assigned to driver $i$};\\
0 & \mbox{if order $j$ is not assigned to driver $i$}.\end{array} \right. \]
which can be solved by standard matching algorithms
\cite{munkres1957algorithms}.

We compute the utility score $\rho_{ij}$ as the TD error defined in \eqref{equ:delta-td-error}, i.e.,
\begin{align*}
    \rho_{ij} = r^j_{order} + \gamma^{\Delta t_{order}} V_{\theta}(s^j_{destination}) - V_{\theta}(s^i_{driver})
\end{align*}
where $r^j_{order}$ denotes the trip fee the driver receives by serving the order $j$, $\Delta t_{order}$ is the estimated trip duration, $s^j_{destination}$ represents the state at the destination of the order $j$.
As TD error $\rho_{ij}$ computes the difference between the expected return of a driver $i$ accepting order $j$ 
and that of the driver staying where she is. Alternatively it can also be seen as the \emph{advantage} of executing the option of picking up the order $j$ compared to the option of no movement.
For each driver such \emph{advantage} is different when pairing with different orders. And our objective is to maximize the total \emph{advantage} collectively for all drivers under constraints.

\begin{figure}[]
\begin{center}
    \hspace*{-0.0in}\adjincludegraphics[width=0.48\textwidth, trim={0 0 0 0}, clip]{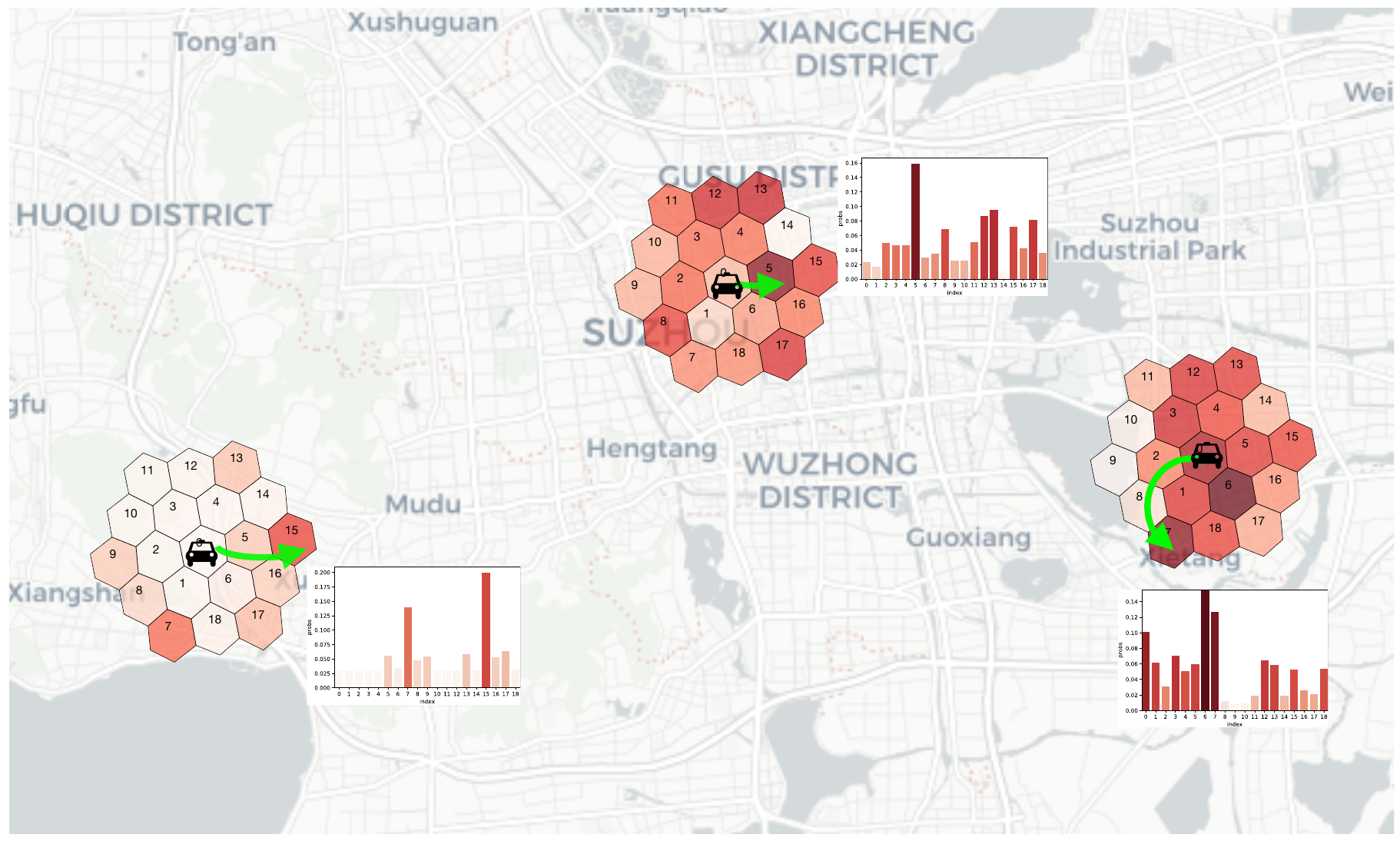}
    \caption{Illustrations of the reposition process by V1D3. The green arrow denotes the reposition direction sampled from the grid value distribution in the vicinity of the vehicle. }
    \label{fig:repo_illustr}
\end{center}
\end{figure}

\subsection{Large-scale Fleet Management} 
\label{sub:value_based_large_scale_fleet_management}

Consider the set of repositioning drivers as $\Ical$ which includes all drivers with idle time exceeding a threshold of C minutes (typically five to ten minutes). For each driver $i \in \Ical$, we consider repositioning the driver to a location selected from a set of candidate destinations $\Ocal_d(s^i)$ given the driver's current state $s^i$.
By doing so we expect to maximize the expected long-term return for the driver, i.e., the value of the destination state. Particularly, we sample the reposition destination with probability proportional to the discounted state value function, e.g.,
\begin{align}
\label{equ:reposition-prob}
    p(s^i_k) \sim
    \frac{e^{\gamma^{\Delta t_{ik}} V_{\theta}(s^i_k)}}{\sum_{j \in \Ocal_d(s^i)} e^{\gamma^{\Delta t_{ij}} V_{\theta}(s^i_j)}}, ~\forall k \in \Ocal_d(s^i)
\end{align}
where $0 < \gamma \leq 1$ is the discount factor and $\Delta t_{ik}$ represents the estimated travel time to the destination $k$. We always include the driver's current location in the candidate set $i \in \Ocal_d(s^i)$. In that case the travel time will be zero $\Delta t_{ii} = 0$ and the state value will not be discounted when computing the sampling probability. That is, the cost of repositioning to a state different from the current is accounted for such that a closer destination with a smaller reposition cost is preferred given the same state value.


\begin{algorithm}
\caption{Unified \textbf{V}alue Learning Framework for \textbf{D}ynamic Order \textbf{D}ispatching and \textbf{D}river Repositioning (\textbf{V1D3})}
\begin{algorithmic}[1]\label{alg:unified-framework}
\STATE Given: the ensemble weight $1 > \omega > 0$, the reposition threshold $C > 0$ (usually chosen between 150 and 300).
\STATE Given: the offline evaluated value function $V_{ope}$.
\label{stp:changing-time-set}
\STATE Compute the set $\Ecal$ containing the changing time points to re-ensemble.
\STATE Initialize the state value network $V$ with random weights $\theta$.
\FOR{the dispatch round $t = 1, 2, \cdots, N$}
  \IF{$t \in \Ecal$}
        \label{stp:ensemble}
        \STATE $\forall s, ~V_{\theta}(s) \leftarrow \omega V_{\theta}(s) + (1 - \omega) V_{ope}^t(s)$.
  \ENDIF
  \label{stp:dispatch}
  \STATE Solve the dispatch problem \eqref{equ:dispatch-obj} given the current value $V_{\theta}$.
  \IF{$t \mod C = 0$}
    \STATE Collect all drivers with idle time exceeding C time steps.
    \STATE Compute the destination distribution \eqref{equ:reposition-prob} for each driver given the current value $V_{\theta}$.
    \label{stp:repo}
    \STATE Reposition each driver stochastically according to the distribution.
  \ENDIF
  \STATE Obtain the system state $\Dcal_D$, $\Dcal_I$ and $\Dcal = \Dcal_D \cup \Dcal_I$.
  \STATE Construct the gradient of the learning objective \eqref{equ:pop-td-error}, i.e., $\nabla L(\Dcal; \theta)$ based on the current system state $\Dcal$.
  \label{stp:sgd}
  \STATE Update the state value network by performing a gradient descent step on $\theta$, e.g.,  $\theta \leftarrow \theta - \alpha \nabla L(\Dcal;\theta)$
\ENDFOR
\RETURN $V$
\end{algorithmic}
\end{algorithm}

\subsection{Unified Value-based Learning Framework} 
\label{sub:unified_value_based_learning_framework}
We combine on-policy learning and offline policy evaluation for both order dispatching and vehicle repositioning in a unified framework presented in Algorithm \ref{alg:unified-framework}.
In particular, at the beginning of each episode we initialize the state value network $V$ with random weights $\theta$, and obtain the offline policy network $V_{ope}$ pretrained using \eqref{equ:ope-vnet}. The training states are augmented by additional time stamp inputs on usually one month of order transactions and driver trajectories collected through the ride-hailing platform.
At Step \ref{stp:changing-time-set} we precompute the set $\Ecal$ of changing time points
by learning a segmentation on the historical aggregated order time series.

In the main loop, for each dispatching round $t$ (usually every 2 seconds), we re-ensemble the value network at Step \ref{stp:ensemble} if $t \in \Ecal$. For a tabular value function this is simply a weighted average between $V_{\theta}$ and $V_{ope}^t$ by traversing each state in the table. For a neural network representation of the value function, this can be done by knowledge distillation \cite{distill2015} with the RHS of \eqref{equ:weighted_ensemble} as the distillation target. The dataset for distillation can be a combination of all states seen so far in the \emph{current} episode and a subsample of historical states.

At Step \ref{stp:dispatch} - \ref{stp:repo} we dispatch outstanding orders and reposition idle drivers based on the current state value network $V_{\theta}$. The dispatch is done by solving the assignment problem \eqref{equ:dispatch-obj} to maximize the total \emph{advantage} of the driver population under constraints. The reposition is only triggered every $C \approx 150$ time steps and only for those drivers with idle time exceeding $C$ time steps.
The ability to scale to a large number of reposition drivers with minimal performance loss is enabled by the following two purposely designed elements of the framework. First of all, the value network is updated for each round of dispatching, reflecting the value changes due to the state transitions of drivers in real time.
Between two reposition actions the value network is updated $C \approx 150$ times which serves as implicit coordinations such that the current reposition computation is aware of the results of the last reposition action.
Secondly, within each reposition computation, stochasticity is added by sampling action from the value distribution \eqref{equ:reposition-prob}. This enhances explorations while avoiding crowding all idle drivers into a single location with the highest value.

Finally, the value network $V_{\theta}$ is updated at Step \ref{stp:sgd} by taking a gradient step of the online learning objective \eqref{equ:pop-td-error}. In practice momentum \cite{nesterov1983}
and adaptive learning rate \cite{adam} can also be used for better learning performance.


\addtolength{\tabcolsep}{3pt}
\begin{table*}
\begin{threeparttable}
  \caption{Comparison with state-of-the-art dispatching algorithms in simulating environments using real-world data from DiDi's ride-hailing platform during both weekdays and weekends in three different cities. The results are averaged from multiple days and the means and variances across days are reported.}
  \label{table:dispatch_results}
  \centering
  \begin{tabular}{c| c| c c c c}
\toprule
City & Environment & Method & Dispatch score  &  Answer rate (\%) \tnote{$\dagger$} & Completion rate (\%) \tnote{$\dagger$}  \\
\hline
\multirow{10}{*}{City A} &
\multirow{5}{*}{Weekday} &
PolarB  & 2498023.82  $\pm$ 12517.26  & +2.8398 $\pm$ 0.3638  & +1.8177 $\pm$ 0.3192  \\
&& Baseline  & 2387008.73  $\pm$ 5429.38 & +0.0000 $\pm$ 0.0000  & +0.0000 $\pm$ 0.0000  \\
&& CVNet  & 2398814.43  $\pm$ 12839.90  & \textbf{+3.7166 $\pm$ 0.3602}  & +0.6548 $\pm$ 0.3540  \\
&& Greedy  & 2350685.21  $\pm$ 5567.51 & -1.2964 $\pm$ 0.0603  & -3.6622 $\pm$ 0.0008  \\
&& V1D3  & \textbf{2509547.65 $\pm$ 8794.37} & +3.0823 $\pm$ 0.0653  & \textbf{+2.0828 $\pm$ 0.0338}  \\
\cline{2-6}
&\multirow{5}{*}{Weekend} &
PolarB  & 2577002.60  $\pm$ 91071.56  & +2.0634 $\pm$ 0.4399  & +0.9494 $\pm$ 0.4347  \\
&& Baseline  & 2487915.88  $\pm$ 77111.26  & +0.0000 $\pm$ 0.0000  & +0.0000 $\pm$ 0.0000  \\
&& CVNet  & 2534253.10  $\pm$ 84285.72  & \textbf{+4.9861 $\pm$ 0.1908}  & \textbf{+1.6428 $\pm$ 0.2126}  \\
&& Greedy  & 2430412.20  $\pm$ 77133.57  & -1.5470 $\pm$ 0.4394  & -4.2193 $\pm$ 0.3719  \\
&& V1D3  & \textbf{2590333.62 $\pm$ 99474.20}  & +2.5222 $\pm$ 0.1956  & +1.3679 $\pm$ 0.1300  \\
\hline
\multirow{10}{*}{City B} &
\multirow{5}{*}{Weekday} &
PolarB  & 1575231.41  $\pm$ 29200.11  & +2.5077 $\pm$ 2.0896  & +1.1372 $\pm$ 1.9432  \\
&& Baseline  & 1498126.49  $\pm$ 12037.66  & +0.0000 $\pm$ 0.0000  & +0.0000 $\pm$ 0.0000  \\
&& CVNet  & 1511983.792 $\pm$ 12331.36  & +2.6405 $\pm$ 0.3073  & +0.2856 $\pm$ 0.2215  \\
&& Greedy  & 1498385.19  $\pm$ 30811.10  & +1.2401 $\pm$ 1.4075  & -1.3727 $\pm$ 1.3386  \\
&& V1D3  & \textbf{1589252.82  $\pm$ 20981.18}  & \textbf{+3.7677 $\pm$ 0.7358}  & \textbf{+2.4352 $\pm$ 0.5846}  \\
\cline{2-6}
&\multirow{5}{*}{Weekend} &
PolarB  & 1436435.90  $\pm$ 52206.43  & +1.3003 $\pm$ 1.4210  & -0.2523 $\pm$ 1.5487  \\
&& Baseline  & 1402633.35  $\pm$ 33007.10  & +0.0000 $\pm$ 0.0000  & +0.0000 $\pm$ 0.0000  \\
&& CVNet & 1407527.12  $\pm$ 38468.35  & \textbf{+2.5140 $\pm$ 1.4626}  & -0.8369 $\pm$ 1.5392  \\
&& Greedy  & 1388862.54  $\pm$ 46301.08  & +0.6618 $\pm$ 0.6337  & -2.3576 $\pm$ 0.9062  \\
&& V1D3  & \textbf{1453191.10  $\pm$ 40822.98}  & +2.4246 $\pm$ 0.2247  & \textbf{+0.8618 $\pm$ 0.2460}  \\
\hline
\multirow{10}{*}{City C} &
\multirow{5}{*}{Weekday} &
PolarB  & 767201.73 $\pm$ 33299.30  & -3.0291 $\pm$ 3.6575  & -3.8274 $\pm$ 3.4695  \\
&& Baseline  & 738083.83 $\pm$ 44261.91  & +0.0000 $\pm$ 0.0000  & +0.0000 $\pm$ 0.0000  \\
&& CVNet  & 744578.48 $\pm$ 42294.09  & \textbf{+6.3528 $\pm$ 0.1955}  & +2.7810 $\pm$ 0.6404  \\
&& Greedy  & 724491.04 $\pm$ 46843.13  & -3.1926 $\pm$ 0.8896  & -5.6701 $\pm$ 0.4511  \\
&& V1D3  &  \textbf{778687.02 $\pm$ 48186.72}  & +4.8733 $\pm$ 0.0938  & \textbf{+2.9925 $\pm$ 0.0934}  \\
\cline{2-6}
&\multirow{5}{*}{Weekend} &
PolarB  & 804656.13 $\pm$ 15354.59  & -1.9825 $\pm$ 2.9749  & -2.8981 $\pm$ 2.9205  \\
&& Baseline  & 764460.73 $\pm$ 4893.10 & +0.0000 $\pm$ 0.0000  & +0.0000 $\pm$ 0.0000  \\
&& CVNet  & 780972.50 $\pm$ 18303.07  & \textbf{+7.0296 $\pm$ 2.4580}  & \textbf{+4.3322 $\pm$ 2.4390}  \\
&& Greedy  & 746729.07 $\pm$ 3357.45 & -4.1320 $\pm$ 0.8392  & -5.8998 $\pm$ 0.5004  \\
&& V1D3  & \textbf{825870.31 $\pm$ 7756.72} & +1.6107 $\pm$ 1.1763  & +0.5496 $\pm$ 0.8569  \\
\bottomrule
  \end{tabular}
\begin{tablenotes}
\item[$\dagger$] The reported numbers are relative improvement computed against the Baseline.
\end{tablenotes}
\end{threeparttable}
\end{table*}
\addtolength{\tabcolsep}{-3pt}

\section{Experiments} 
\label{sec:experiments}

In this section we conduct extensive experiments to evaluate the
effectiveness of our proposed method V1D3 on both tasks.
Additionally, to gain more insights into the internal mechanisms of V1D3,
we empirically analyze the value response curve of V1D3 and demonstrate its online fast adaptation capability.
The details of the real-world datasets used in the experiments including a public sample released for KDD Cup 2020 can be found in Appendix \ref{apx:datasets}.

\subsection{Dispatching Results} 
\label{sub:order_dispatching_experiments}

We compare V1D3 with strong baseline and state-of-the-art dispatching algorithms as follows.

\begin{itemize}[leftmargin=*]
    \item \textbf{Baseline}. A myopic dispatching method used by \cite{tang2019vnet,xu2018large} in the comparisons. The method maximizes the current batch rate of dispatching by minimizing the total distance between all assigned driver-order pairs in the matching problem \eqref{equ:dispatch-obj}.
    \item \textbf{CVNet}. An offline value-based method proposed by \cite{tang2019vnet}. The value function is represented by a neural network trained to minimize the TD error on the historical driver trajectories.
    The learning method uses additional techniques such as hierarchical coarse-coded embedding and Lipschitz regularization for better generalization. The method achieves state-of-the-art performance among offline models in both simulations and online AB testing.
    \item \textbf{Greedy} A myopic and greedy method that dispatches according to the order given by sorting driver-request pairs by the amount of revenue the driver receives after completing the request.
    \item \textbf{Polar Bear (PolarB)}.
    The method that received the \textbf{first prize} in the order dispatching task of KDD Cup 2020 RL track competition. The value function is only trained using online samples without the offline training ensemble component.
\end{itemize}

We run each algorithm in simulation environments constructed from real-world ride-hailing data. The simulation data is collected from DiDi's platform in three different cities and on two weekdays and two weekend days. Each simulation replays a full day's ride-hailing transactions with real passenger requests and drivers' online and offline records. In each dispatching round the simulator sends the current outstanding order requests and idle drivers to the dispatching algorithm to compute the dispatch results which are then used to advance the state by the simulator.
The results of the simulations are presented in Table~\ref{table:dispatch_results}. Three main performance metrics are reported for comparisons, of which the dispatch score represents the total income of all drivers and the answer and completion rate measure the user experience of the passengers from making requests to trip completion.

We first notice the two learning-based methods, PolarB and CVNet,
both of which perform well compared to the Baseline achieving an average of $4\%$ and $1\%$ improvement of dispatch score respectively. While the online learning method PolarB shows an advantage over CVNet on dispatch score, CVNet improves the answer and completion rate against PolarB across cities and days. We conjecture that this is due to the larger size of the dataset accessible for offline training which helps CVNet learn a smoother value function that distributes the drivers more evenly across different geographical regions. Both methods exhibit better performance when there exists a larger gap between supply and demand, e.g., PolarB and CVNet improving dispatch score by $>5\%$ and $>2\%$, respectively, in City C which is supply constrained with the largest gap between supply and demand among all three cities.

The best performing method of all is V1D3, which demonstrates a consistent advantage compared to other methods across cities and days, improving the total driver income in City C, for example, by as much as $8\%$ against the Baseline and maintains a $5\%$ lift in most other cases.
In particular, we note that V1D3 combines the advantages of both PolarB and CVNet, outperforming both methods in dispatch score and achieving comparable if not better performance as CVNet in both the answer rate and the completion rate.
Compared to PolarB, V1D3 further increases the dispatch score by as much as $2.64\%$, which is on top of PolarB's averaged $4\%$ improvement over the Baseline.
Besides, V1D3 achieves consistent and significant improvement in both answer and completion rate across all segments.
Overall the results demonstrate the benefits of V1D3 combining online and offline learning in a single framework which improves significantly both the total driver income (dispatch score) and user experience (the answer and completion rate) for the ride-hailing platform.




\begin{figure*}[]
\centering
        \begin{subfigure}[b]{0.49\textwidth}
                \centering
                \includegraphics[width=\linewidth]{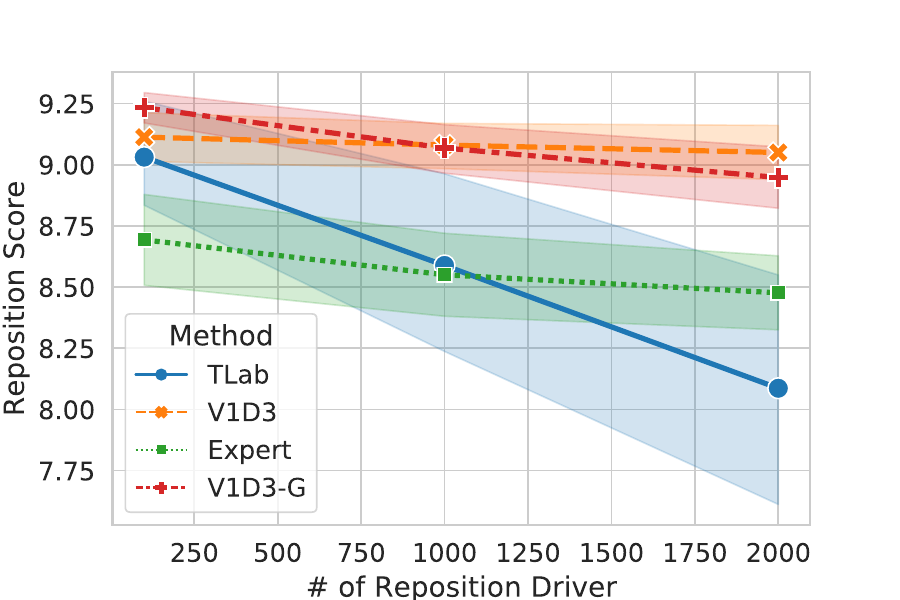}
                \caption{}
                \label{fig:repo}
        \end{subfigure}%
        \vspace{2pt}
        \begin{subfigure}[b]{0.49\textwidth}
                \centering
                \includegraphics[width=\linewidth]{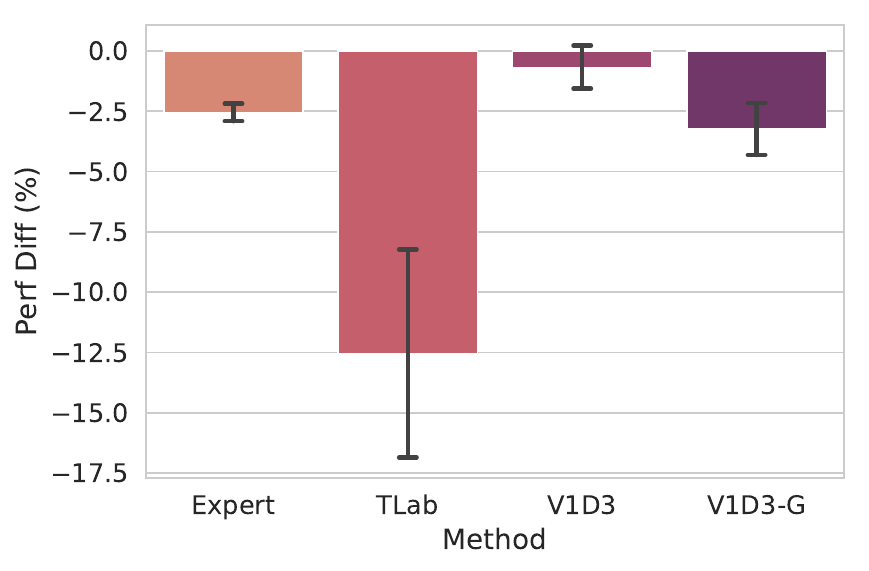}
                \caption{}
                \label{fig:repo_perfdrop}
        \end{subfigure}%
        \caption{
          (a). The drivers managed by V1D3 achieve the highest total income rate (reposition score), surpassing the top solution TLab in KDD Cup 2020 and the human Expert policy.
          (b). The relative difference (\%) of reposition score between N = 100 and 2000 with error bar denoting the standard deviation across 5 days. Here V1D3 demonstrates a robust high performance, within <0.7\% variation, as the managed fleet size N increases 20x.
        }
    \label{fig:repo_analysis}
\end{figure*}

\subsection{Reposition Results} 
\label{sub:reposition_results}
In this subsection we demonstrate the effectiveness of V1D3 as a central repositioning agent managing the idle movements of a fleet of vehicles. We compare with the following methods.
\begin{itemize}[leftmargin=*]
    \item \textbf{TLab}.
    The offline method by \cite{tlab2021} that received the \textbf{first prize} in the vehicle repositioning task of KDD Cup 2020 RL track competition. A single agent deep reinforcement learning approach with a global pruned action space is used. The Q function accepts not only the local information of the managed vehicle as the state but also the global information in the whole area, i.e., the real-time distribution of orders, vehicles and rewards. During inference the reposition action is selected in a greedy manner w.r.t. Q values.
    \item \textbf{Expert}. A human expert policy extracted from the historical idle driver transition data. Millions of idle driver trajectories are analyzed to estimate the transition probability matrix of each origin-destination pair of grids at each given time slice. During inference the reposition destination of a given vehicle is sampled according to the transition matrix.
    \item \textbf{V1D3-G}. A deterministic (greedy) variant of V1D3. During inference instead of sampling the action according to \eqref{equ:reposition-prob} the grid with the highest value is selected with random tie breaking.
\end{itemize}

We use mean income rate, or reposition score, as the metric of the reposition performance. It is computed as the mean of each managed vehicle's total income per unit online time.
To evaluate the algorithm performance we use the same simulator powered by the real-world dataset from the dispatching experiments, in which each managed vehicle acts as an autonomous agent fully controlled by the reposition algorithm. The reposition action is triggered every 150 dispatching rounds (C = 150 in Algorithm~\ref{alg:unified-framework}), i.e., 5 minutes between every two reposition action.
Vehicles idling for less than 5 minutes will not qualify for the reposition round and will stay where they are until dispatched or repositioned.

One of the main challenge in large-scale fleet management is to maintain the high income rate of each individual vehicle as the fleet size increases. To that end the reposition algorithm needs to coordinate the controlled agents and automatically re-balance against dynamic demands, e.g., not directing too many or too few vehicles to the same destination.
Hence the key question we are trying to answer in the experiments is whether the reposition algorithm can still maintain high performance when the size of the managed fleet, denoted as N, increases to a level that takes a significant portion of the total driver population.


\begin{figure*}[h!]
\centering
        \begin{subfigure}[b]{0.24\textwidth}
                \centering
                \includegraphics[width=\linewidth]{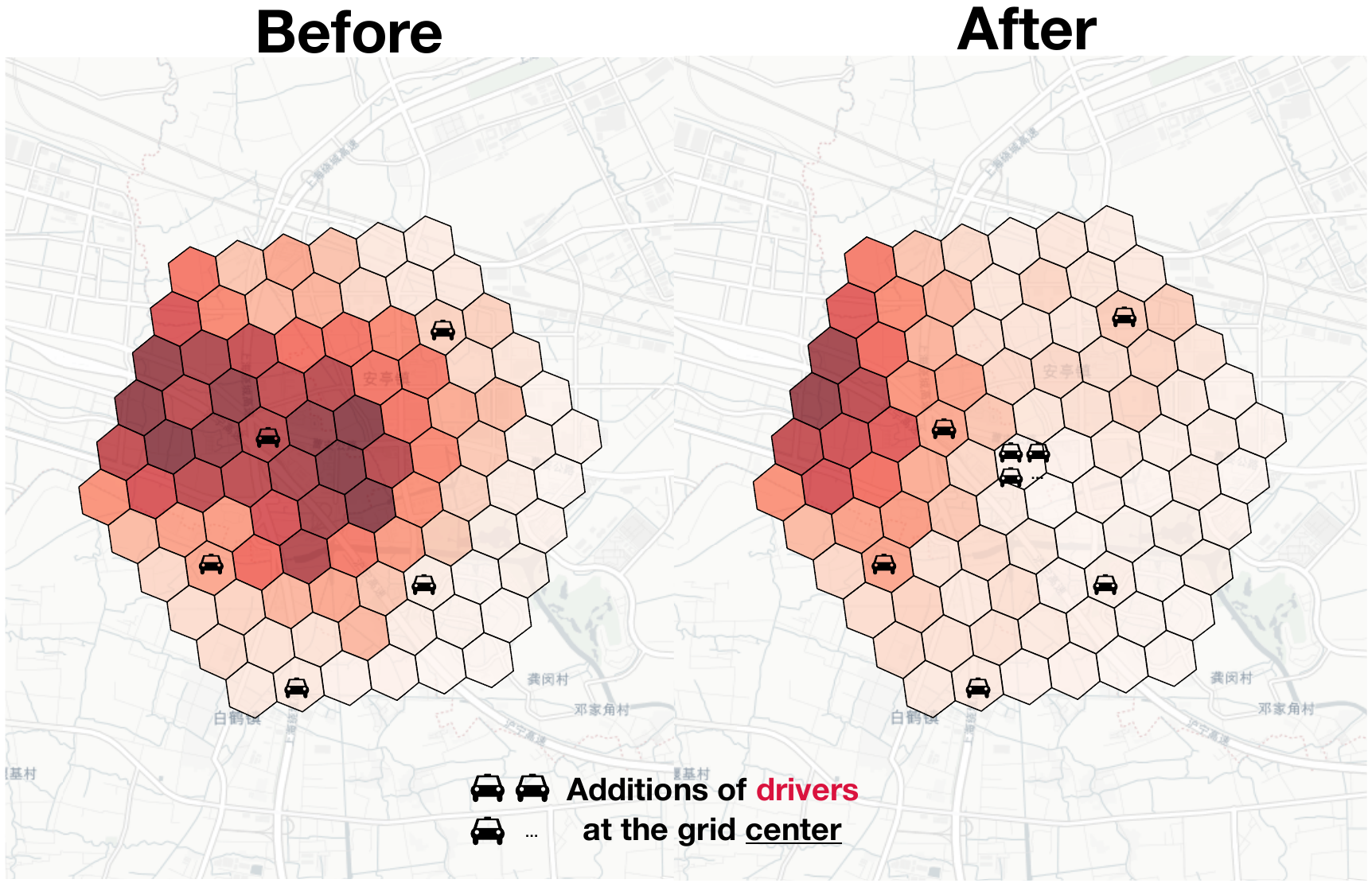}
                \caption{}
                \label{fig:adding_drivers}
        \end{subfigure}%
        \vspace{6pt}
        \begin{subfigure}[b]{0.24\textwidth}
                \centering
                \includegraphics[width=\linewidth]{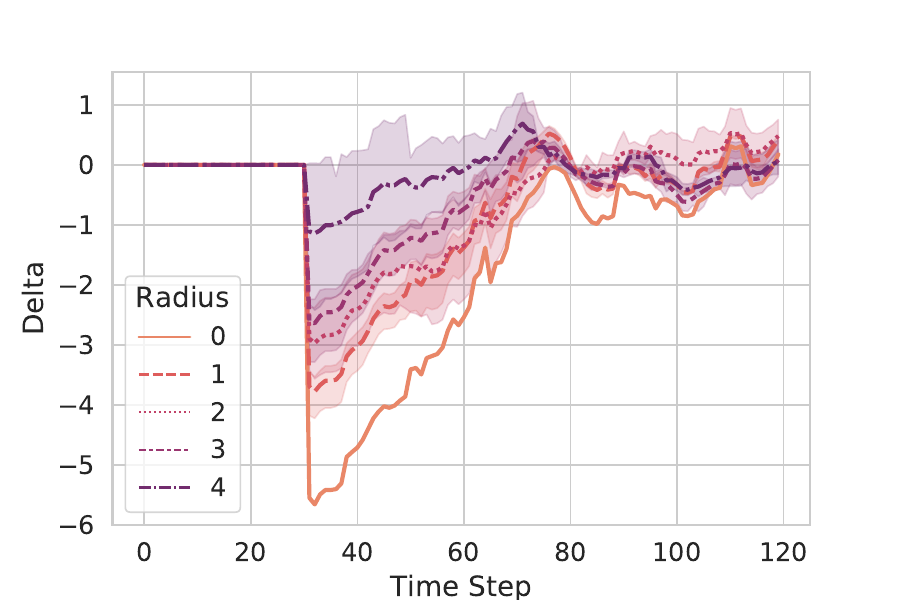}
                \caption{}
                \label{fig:add_drivers_lines}
        \end{subfigure}%
        \begin{subfigure}[b]{0.24\textwidth}
                \centering
                \includegraphics[width=\linewidth]{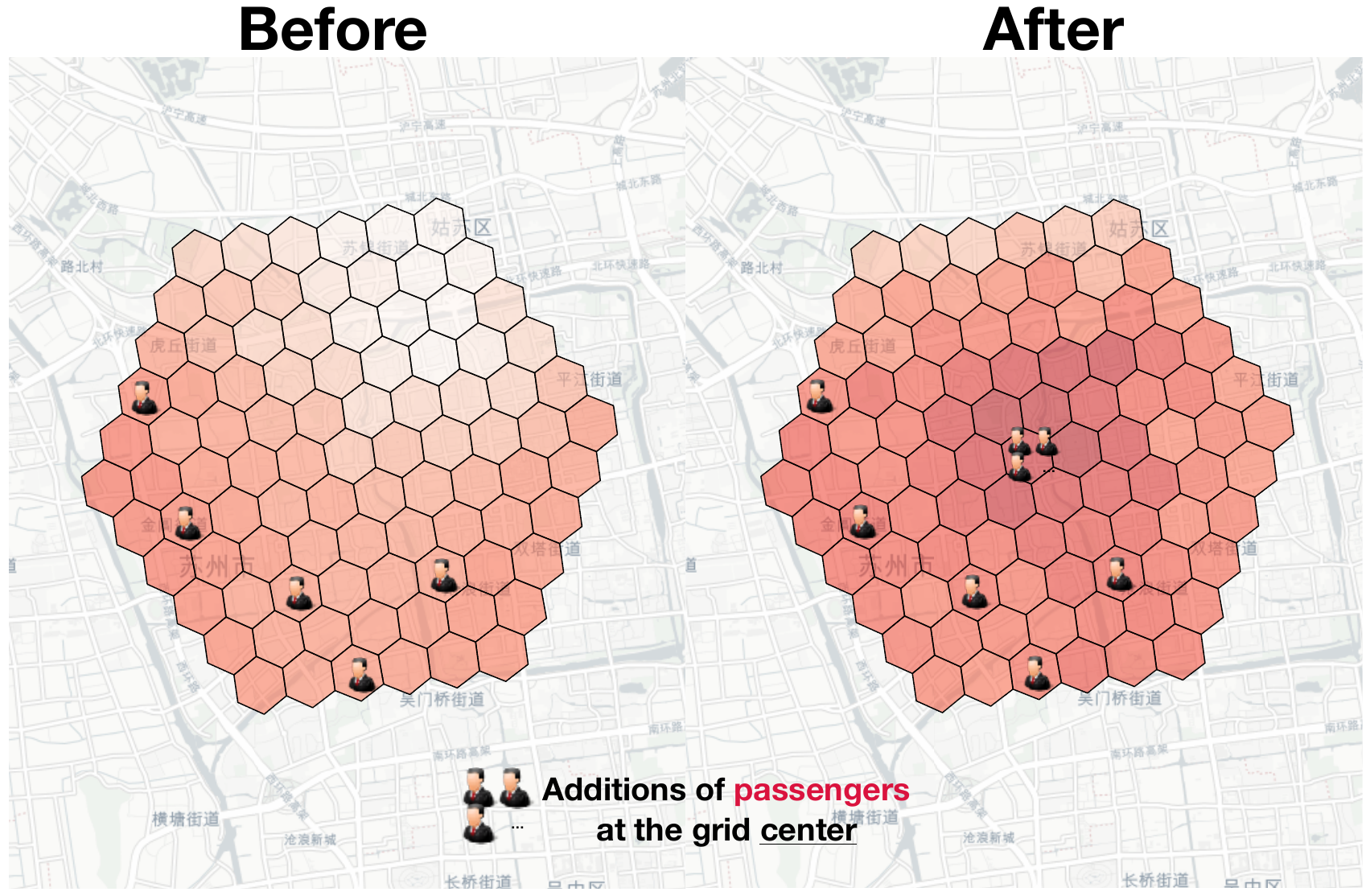}
                \caption{}
                \label{fig:adding_orders}
        \end{subfigure}
        \begin{subfigure}[b]{0.24\textwidth}
                \centering
                \includegraphics[width=\linewidth]{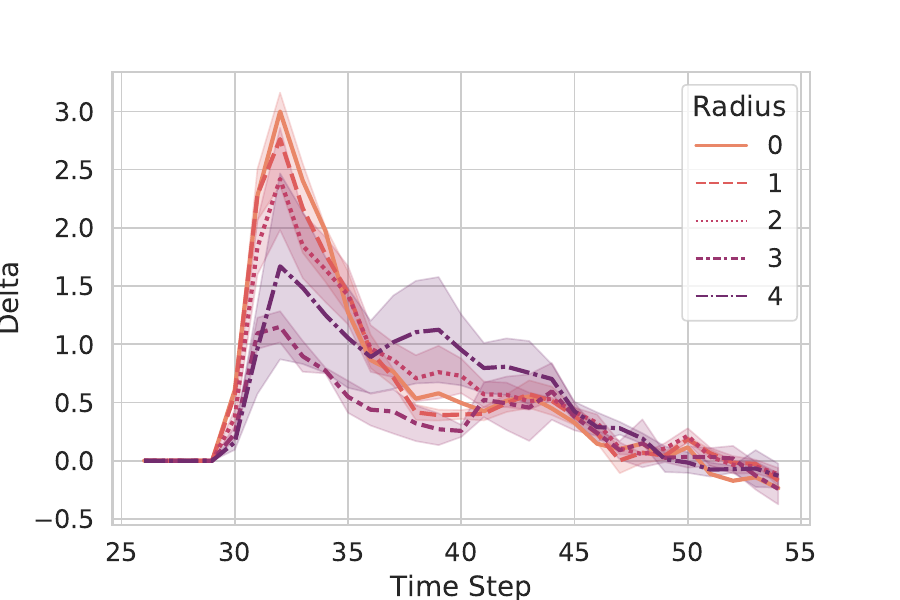}
                \caption{}
                \label{fig:add_orders_lines}
        \end{subfigure}
        \caption{
          (a) and (c). Value distribution before and after additions of new drivers (orders) at the grid center;
          (b) and (d). Each color represents the value distributions of the sets of cells at different radius to the center, e.g., radius 0 denotes the grid center.
          The values decrease (increase) in response to new additional supplies (demands) as desired. The magnitude of the response gradually diminishes as time elapses and as the distance to the grid center increases.
        }
    \label{fig:perf_analysis}
\end{figure*}
The results are presented in Figure~\ref{fig:repo}. Each algorithm is evaluated using real-world data on 5 different dates including both weekdays and weekends. And the same simulations are repeated for $N = 100, 1000, 2000$.
For each N we report the mean (the marker) and the variance (the shaded area) of the reposition score across 5 dates. First we note that when $N$ is small (100) all three learning-based methods outperform the human expert policy by a significant margin, e.g., more than 6\% improvement achieved by the best performer V1D3-G in this case. As N increases, however, the performance of TLab deteriorates dramatically, as much as over 15\%, as shown in Figure~\ref{fig:repo_perfdrop}. On the other hand, V1D3 achieves
a remarkable 12\% improvement on the average income rate compared to TLab as N increases to 2000,
and outperforms consistently with great margin both TLab and the human Expert in all cases.
Note that N=2000 is more than 20\% of the vehicle population in the simulations.
The results show that the proper use of stochastic policy helps improve the robustness to the increase of N (V1D3 vs V1D3-G), and that online learning facilitates effective coordination among managed vehicles and help achieve a stunning performance scalability and robustness when N increases in both deterministic and stochastic cases, e.g., V1D3 outperforms the TLab by 15x and the human Expert policy by nearly 4x as shown in Figure~\ref{fig:repo_perfdrop}.




\subsection{Performance Analysis} 
\label{sub:perf_analysis}
In this subsection we take a deeper look at the online learning component of V1D3.
It enables V1D3 to quickly adapt to any new changes occurring in the online environment and is key to the superior performance of V1D3 in both order dispatching and driver repositioning.
In the following experiments we empirically support this observation by simulating the occurrences of irregular events and visualizing the changes of the values in response to those events. The results are presented in Figure~\ref{fig:perf_analysis}.

\subsubsection{The additions of new drivers} 
\label{ssub:the_additions_of_new_drivers}

This illustrates the real-world cases where drivers (or supplies) become available at the neighborhood of one location, either by completing previous orders or by going online from offline, e.g., at the airport or the busy downtown district.
To visualize the adaptive response of V1D3 to such events, we run the full simulation with new drivers added to the center of the grid at 30th time step, and record the set of grid value time series through the simulation.
As a comparison, we run the same simulation again but without the alterations. The differences between these two sets of values are plotted in Figure~\ref{fig:add_drivers_lines}. It can be seen that in response to the new additional supplies at time step 30, V1D3 quickly decreases the values of the grid, as intended, discouraging more drivers being sent to the location.
The strongest response is observed at the grid center (Radius 0), where the new supplies are created.
Moreover, the nearby cells also see a decrease of the values proportional to the distance from the grid center. This ensures a geographically smooth value response thanks to the coarse-coded representation of the state \cite{tang2019vnet}.
Finally, as time elapses, the additional drivers are dispatched and the state of the system gradually returns back to normal with the value delta decreasing to zero.


\subsubsection{The additions of new orders} 
\label{ssub:the_additions_of_new_orders}

This corresponds to the real-world cases where demands (or passenger requests) surge at a specific location possibly due to some large events like concerts.
To simulate such event
an order generation process is implemented with 10 new orders created every 8 seconds at the grid center. The process lasts for 4 minutes with a total of 300 additional orders created.
To visualize the dynamic response of V1D3 to such event, again two sets of value time series are recorded in two separate simulation runs, one with the additions of new orders and the other without. Their differences are plotted in Figure~\ref{fig:add_orders_lines}.
Note that value difference (delta) increases from zero at time step 30 and peaks at time step 32 where the order generation process ends. There is a two-minutes gap between two adjacent time steps in Figure~\ref{fig:add_orders_lines} so the generation process lasts for 2 time steps.
Again the increase of values at the presence of new additional orders is observed, as desired, such that it encourages more drivers being sent over to satisfy the increased demands. And as time elapses the new orders are being fulfilled and the state of the system returns back to normal with the value delta decreasing to zero.
The two sets of results presented in Figure~\ref{fig:perf_analysis} together demonstrate that V1D3 can quickly adapt to the variations of both supplies (drivers) and demands (orders) with the dynamic responses of the values, which in turn act to balance supplies and demands through dispatching and repositioning.
This makes it possible for V1D3 to achieve superior performance in improving both driver efficiency and passenger answer rate (user experiences) in the ride-hailing markets.





\section{Conclusions} 
\label{sec:conlude}
In this paper, we consider the problems of dispatching and repositioning, the two major operations in ride-hailing platforms. We propose the use of a centralized value function as the basis for learning and optimization to capture the interactions between the two tasks. To that end, we propose a novel value ensemble approach that enables fast online learning and large-scale offline training. The online learning is based on a novel population-based Temporal Difference objective derived from the on-policy value iteration and allows the value function to adapt quickly to new experiences from the platform's online transactions. Finally, we unify these advancements in a value-based learning framework V1D3 for both tasks.
Through extensive experiments based on real-world datasets, V1D3 shows considerably improvements over other recently proposed methods on both tasks, outperforming the first prize winners of both dispatching and repositioning tracks, respectively, in the KDD Cup 2020 RL competition, and achieving state-of-the-art results on improving both total driver income and user experience related metrics. Going forward this work provides a strong algorithmic foundation for real-world large-scale deployment which we are actively work on.
Hundreds of millions of passengers and drivers stand to enjoy the improvements brought by our new algorithm.

\bibliographystyle{abbrv}
\balance
\bibliography{vnet}

\clearpage
\appendix

\begin{figure}[]
\begin{center}
    \hspace*{-0.0in}\adjincludegraphics[width=0.48\textwidth, trim={0 0 0 0}, clip]{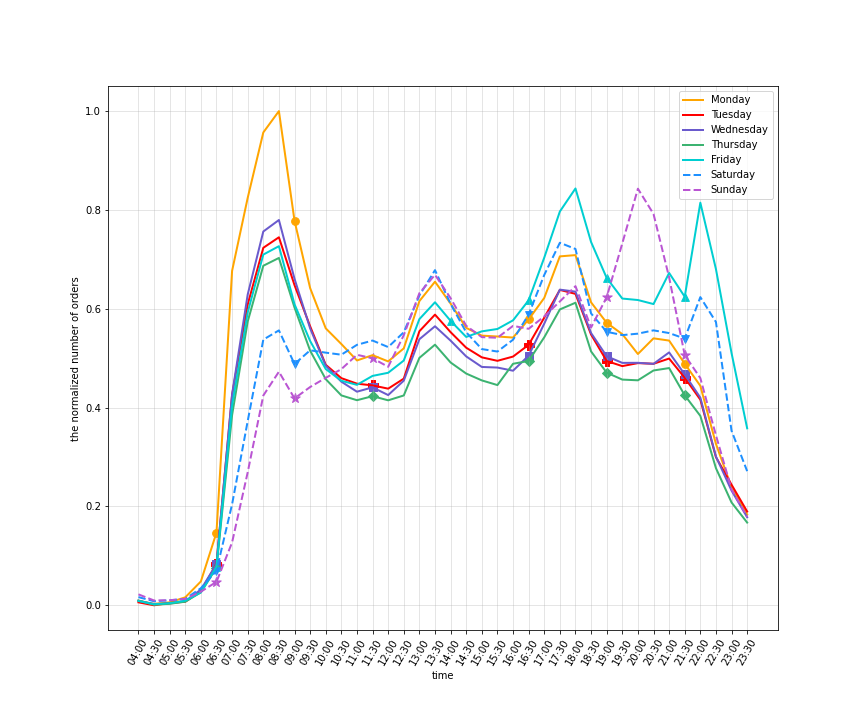}
    \caption{Illustrations of the order time series segmentation. Solid (dashed) lines represent the trend of changes in the number of orders on weekdays (weekends). Markers are change points detected by the offline change point detection algorithm.}
    \label{fig:changing_time}
\end{center}
\end{figure}

\begin{algorithm}
\caption{Regularized Offline Policy Evaluation (OPE)}
\begin{algorithmic}[1]\label{alg:dpe}
\STATE Given:
historical driver trajectories $\{(s_{i,0}, o_{i,0}, r_{i,1}, s_{i,1}, o_{i,1}, r_{i,2}, ..., r_{i,T_i}, s_{i,T_i})\}_{i \in \mathcal{H}}$ collected by executing a (unknown) policy $\pi$ in the environment.
\STATE Given: the learning rate $\alpha$, $n$ cerebellar quantization functions $\{q_1, ..., q_n\}$, and $\lambda, N, A, m, g(\cdot), \gamma$ which are, respectively, regularization parameter, max iterations, embedding memory size, embedding dimension, memory mapping function, discount factor.
\STATE Compute training data from the driver trajectories as a set of (state, reward, next state) tuples, e.g., $\{(s_{i,t}, R_{i,t}, s_{i,t+k_{i,t}})\}_{i \in \mathcal{H}, t=0,...,T_i}$ where $k_{i, t}$ is the duration of the trip.
\STATE Initialize the embedding weights $\rho^M$.
\STATE Initialize the state value network $V_{ope}(\cdot|\rho)$ with random weights $\rho$.
\FOR{$\kappa = 1, 2, \cdots, N$}
  \STATE Sample a random mini-batch $\Dcal_{\Bcal} := \{(s_{i,t}, R_{i,t}, s_{i,t+k_{i,t}})\}_{i\in\Bcal}$ from the training data.
  \STATE Embed the states $s_{i, t}$ and $s_{i, t+k_{i,t}}$ using the quantization functions $\{q_1, ..., q_n\}$ and the memory mapping function $g(\cdot)$, e.g., $s \leftarrow c(s)^T \rho^M / n$ where the activation vector $c(s)$ is initialized to 0 and iteratively adding 1 to the $g(q_i(s))$-th entry of $c(s)$ for $i = 1, ..., n$.
  \STATE Evaluate the loss function \eqref{equ:ope-vnet}, e.g.,
  $L_{ope}(\Dcal_{\Bcal}; \rho) = \sum_{i\in\Bcal}(\frac{R_{i,t}(\gamma^{k_{i,t}} - 1)}{k_{i,t}(\gamma - 1)} + \gamma^{k_{i,t}} V_{ope}(s_{i, t+k_{i,t}}) - V_{ope}(s_{i, t}))^2 + \lambda \cdot L_{reg}(\rho)$
  \STATE Update the state value network $V_{ope}(\cdot|\rho)$ with gradient descent, e.g., $\rho \leftarrow \rho - \alpha \nabla L_{ope}(\Dcal_{\Bcal}; \rho)$
\ENDFOR
\RETURN $V_{ope}$
\end{algorithmic}
\end{algorithm}

\section{Implementation Details}
\label{apx:impl}

\subsection{Hyperparameters of V1D3} 
\label{apx:hyperparameters_of_v1d3}
We tune all the hyperparameters by running the simulation on one of the weekdays in City A.
The parameters are then fixed and shared in all experiments reported in the paper.
Specifically, the ensemble weight $\omega$ is chosen to be 0.2, putting more weight on the offline policy $V_{ope}$.
The reposition threshold $C$ is 150 which is 5 minutes in real time. We use a discount factor $\gamma$ of 0.9 to evaluate the online objective \eqref{equ:pop-td-error}. The online learning rate is fixed at $0.025$.


\subsection{Order Time Series Segmentations} 
\label{apx:ecal}

We apply the algorithm described in \cite{TRUONG2020107299} to the historical aggregated order time series to determine the changing point set $\Ecal$ when the value network is re-ensembled with the corresponding time snapshot of the offline policy network.
Specifically, we first count the number of orders at each fixed time interval from 4:00 to 24:00 according to the order data collected one week before the simulation date with the fixed time interval set to 30 minutes in practice. Then we pass the number of change points K and order time series on half-hour basis into the dynamic programming method \cite{TRUONG2020107299} to detect the changing points.
As a result, the approach detects K change points and divides order time series into K+1 segments.
K is set to 5 in this work based on the tuning procedures described above.
As is illustrated in the Figure \ref{fig:changing_time}, there are seven lines, each showing the order time series on each day of a certain week in City A.  The solid and dashed lines are used to distinguish two different urban travel demand patterns of weekdays and weekends.
Moreover, markers on the lines represent change points detected by algorithms.
We can see that there are strong patterns in the time series as well as the detected segmentations. The first set of changing points, for example, are always detected as 6:30, which is corresponding to the start of morning peak.
The patterns in weekdays are usually different from those on weekends.
For instance, the morning rush hours in most weekdays usually end at 11:30, while morning peaks on weekends are shorter and usually end at 9:00.


\subsection{Offline Policy Evaluation}
\label{apx:ope}
The main procedure of OPE is presented in Algorithm \ref{alg:dpe}.
We use the same discount factor $\gamma = 0.9$ in OPE.
Following \cite{tang2019vnet}, we employ cerebellar embedding to obtain a distributed state representation, and Lipschitz regularization $L_{reg}$ for a robust and stable training and state value dynamics.
The cerebellar embedding defines multiple quantization functions, each of which maps continuous inputs to embedding vectors which together form the memory of the CMACs.
The Lipschitz regularization aims to control the global Lipschitz constant of the neural network that represents the value function to be evaluated. This can be decomposed to regularizing the Lipschitz constant of each individual layer in the network. For the cerebellar embedding layer, it is equivalent to minimizing the infinity norm of the embedding matrix.
In this work, we use 3 quantization functions and a memory size $A$ of 20000 for the cerebellar embedding. The embedding dimension $m$ is chosen to be 50.
Following the cerebellar embedding layer are fully connected layers having [32, 128, 32] hidden units with ReLU activations.
To evaluate the policy we apply Adam optimizer with a constant step size $3e^{-4}$ and the Lipschitz regularization parameter $\lambda$ as $1e^{-4}$.

\section{Simulation Datasets}
\label{apx:datasets}
The anonymized dataset used to evaluate the algorithm consists of four parts: the driver trajectory data, the trip request data, the transition data, and the order cancellation data. Each record in the trajectory data is a tuple of the driver id, the order id, the current timestamp, and the current latitude and longitude of the driver. The trip request data describes order the details for each order id, including the ride start and stop time, the pickup and drop-off locations, and the reward units. The transition data is obtained by training on the drivers' idle behavior in the real world, with the transition probability for each pair of origin and destination grids for each hour of the day. Finally, the order cancellation data dictates the behavior of order cancellations with respect to the pick-up distances, which captures the nonlinear response of the cancellation probability.
A sample of this dataset is released in KDD Cup 2020 RL competition and can be accessed through the Gaia Initiative (\url{https://outreach.didichuxing.com/research/opendata/en/}).
The competition evaluation platform powered by this dataset is also open for public and can be accessed at \url{https://outreach.didichuxing.com/Simulation/}.

\end{document}